%% file: thewlis16flow.tex
\documentclass{bmvc2k}
\usepackage{graphicx}
\usepackage{amsmath,amssymb}
\usepackage{bm}
\usepackage{color}
\usepackage{xspace}
\usepackage{epsfig}
\usepackage{adjustbox}
\usepackage{changepage}
\usepackage{caption}

\def\ie{\emph{i.e}\bmvaOneDot}

\def\etal{\emph{et al}\bmvaOneDot}

\newcommand{\bi}{\mathbf{i}}
\newcommand{\bk}{\mathbf{k}}

\newcommand{\bp}{\mathbf{p}}
\newcommand{\bq}{\mathbf{q}}

\newcommand{\bw}{\mathbf{w}}
\newcommand{\bx}{\mathbf{x}}

\newcommand{\bdelta}{\bm\delta}
\newcommand{\bepsilon}{\bm\epsilon}

\renewcommand{\paragraph}[1]{\par\smallskip\noindent{\bf #1}}

\title{Fully-Trainable Deep Matching}
\addauthor{James Thewlis}{jdt@robots.ox.ac.uk}{1}
\addauthor{Shuai Zheng}{shuai.zheng@eng.ox.ac.uk}{1}
\addauthor{Philip H. S. Torr}{philip.torr@eng.ox.ac.uk}{1}
\addauthor{Andrea Vedaldi}{vedaldi@robots.ox.ac.uk}{1}
\addinstitution{
 Department of Engineering Science\\
 University of Oxford\\
 Oxford, UK
}
\runninghead{Thewlis~\emph{et al.}}{Fully-Trainable Deep Matching}

\begin{document}

\maketitle
\begin{abstract}
Deep Matching (DM) is a popular high-quality method for quasi-dense image matching. Despite its name, however, the original DM formulation does not yield a deep neural network that can be trained end-to-end via backpropagation. In this paper, we remove this limitation by rewriting the complete DM algorithm as a convolutional neural network. This results in a novel deep architecture for image matching that involves a number of new layer types and that, similar to recent networks for image segmentation, has a $U$-topology. We demonstrate the utility of the approach by improving the performance of DM by learning it end-to-end on an image matching task.
\end{abstract}

\section{Introduction}

\emph{Deep Matching} (DM)~\cite{Revaud_ijcv2015} is one of the most popular methods for establishing quasi-dense correspondences between images. An important application of DM is optical flow, where it is used for finding an initial set of image correspondences, which are then interpolated and refined by local optimisation.

The reason for the popularity of DM is the quality of the matches that it can extract. However, there is an important drawback: DM, as originally introduced in~\cite{Revaud_ijcv2015}, is in fact \emph{not} a deep neural network and does not support training via back-propagation. In order to sidestep this limitation, several authors have recently proposed alternative Convolutional Neural Networks (CNN) architectures for dense image matching (Sect.~\ref{s:related}). However, while several of these trainable models obtain excellent results, they are not necessarily superior to the handcrafted DM architecture in term of performance.

The quality of the matches established by DM demonstrates the strength of the DM architecture compared to alternatives. Thus, a natural question is whether it is possible to obtain the best of both worlds, and construct a trainable CNN architecture which is equivalent to DM. The main contribution of this paper is to carry out such a construction.

\begin{figure}[t]
\begin{center}
\includegraphics[width=\textwidth]{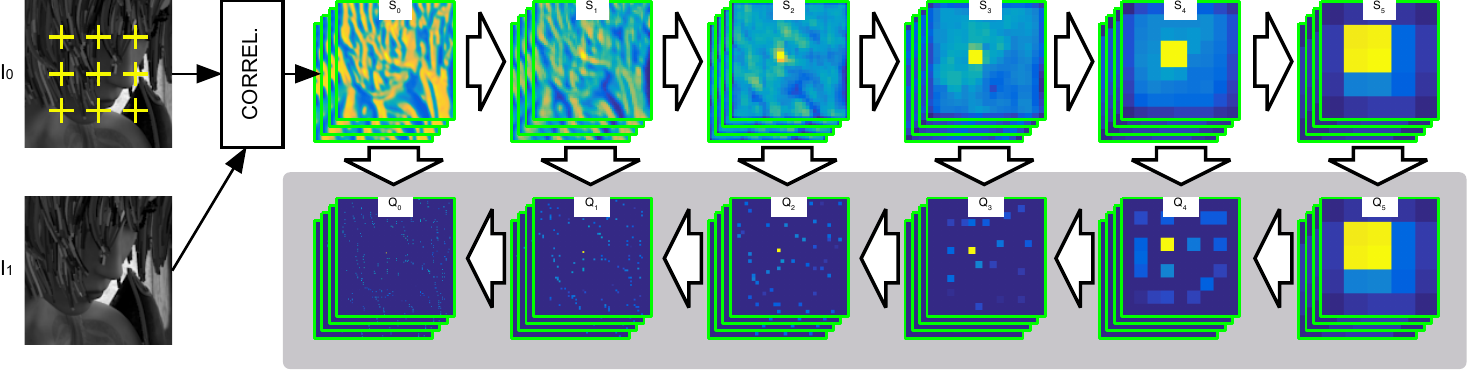}
\end{center}
\caption{\textbf{Fully-Trainable Deep Matching.} Deep matching starts by correlating small patches $\bp$ in the reference image $I_0$ (crosses) with all the patches $\bq$ in the target image $I_1$, producing a 4D score map $S_0$ (slices $S_0(\cdot|\bp)$ for varying $\bp$ are shown); then, it computes coarser but less ambiguous maps $S_1,\dots,S_L$. In this paper, we formulate the reverse process, reconstructing high resolution matches from coarser ones, as a sequence of \emph{reverse convolutional operators}, producing scores $Q_L,\dots,Q_0$ (shaded area). The result is a deep convolutional network with $U$-architecture that can be trained using backpropagation.}\label{f:splash}
\end{figure}

In more detail, DM comprises two stages (Fig.~\ref{f:splash}): In the first stage, DM computes a sequence of increasingly coarse match score maps, integrating information from progressively larger image neighbourhoods in order to remove local match ambiguities. In the second stage, the coarse information is propagated in the reverse direction, resolving ambiguities in the higher-resolution score maps. While the first stage was formulated as a CNN in~\cite{Revaud_ijcv2015}, the second stage was given as a recursive decoding algorithm. In Sect.~\ref{s:method}, we show that this recursive algorithm is equivalent to dynamic programming and that it can be implemented instead by a sequence of new \emph{convolutional operators, that reverse the ones in the first stage of DM}.

The resulting CNN architecture (Fig.~\ref{f:arch}), which is \emph{numerically equivalent to the original DM}, has a $U$-topology, as popularized in image segmentation~\cite{Ronneberger/miccai2015}, and supports backpropagation. Combined with a structured-output loss  (Sect.~\ref{s:training}), this allows us to perform end-to-end learning of the DM parameters, improving its performance (Sect.~\ref{s:experiments}). Our findings and further potential advantages of the approach are discussed in Sect.~\ref{s:summary}.

\input{relatedworks}
\input{approach}
\input{experiments2}

\section{Summary}\label{s:summary}

In this paper, we have shown that the \emph{complete} DM algorithm can be equivalently rewritten as a CNN with a $U$-topology, involving a number of new CNN layers. This allows to learn end-to-end the parameters of DM using backpropagation, including the CNN filters that extract the patch descriptors, robustly improving the quality of the correspondence extracted in a number of different datasets.

Once formulated as a modular CNN, components of DM can be easily replaced with new ones. For instance, the max pooling and unpooling units could be substituted with soft versions, resulting in denser score maps, which could result in easier training and in the ability of better expressing the confidence of dense matches. We are currently exploring a number of such extensions.

For the problem of optical flow estimation, it is still required to have EpicFlow as a post-processing step. This type of two-stage approach results a suboptimal solution. In particular, the parameters of EpicFlow are not optimized by end-to-end training with our DM. We would like to explore a solution that allows end-to-end optical flow estimation.

\paragraph{Acknowledgements.}
{This work was supported by the AIMS CDT (EPSRC EP/L015897/1) and grants EPSRC EP/N019474/1, EPSRC EP/I001107/2, ERC 321162-HELIOS, and ERC 677195-IDIU. We gratefully acknowledge GPU donations from NVIDIA.}

\bibliography{matchingflow}
\end{document}

%% file: relatedworks.tex
\subsection{Related Work}\label{s:related}


The key reason for the success of CNNs in many computer vision applications is the ability to learn complex systems end-to-end instead of hand-crafting individual components. A number of recent works have applied CNN-based systems to pixel-wise labeling problems such as stereo matching and optical flow. In particular, Fischer~\etal~\cite{Fischer/iccv2015} have shown it is possible to train a fully convolutional network for optical flow. {\v Z}bontar~\etal~\cite{Zbontar/arxiv2015} trained a CNN for stereo matching by using a refined stereo matching cost. Zagoruyko and Komodakis~\cite{Zagoruyko/cvpr2015} and Han~\etal~\cite{Han/cvpr2015} have demonstrated learning local image description through a CNN.

Optical flow estimation was tackled mostly by variational approaches~\cite{Memin/tip1998,Brox/eccv2004,Wedel/iccv2009} since the work of Horn and Schunk~\cite{Horn/ai1981}. Brox and Malik~\cite{Brox/pami2011} developed a system that integrates descriptor matching with a variational approach. Recently, leading optical flow approaches such as DeepMatching~\cite{Weinzaepfel/iccv2013,Revaud_ijcv2015} demonstrated a CNN-like system where feature information is aggregated from fine to coarse using sparse convolutions and max-pooling. However, this approach does not perform learning and all parameters are hand-tuned. EpicFlow~\cite{Revaud/cvpr2015} has focused on refining the sparse matches from DM using a variational method that incorporates edge information. Fischer~\etal~\cite{Fischer/iccv2015} trained a fully convolutional network FlowNet for optical flow prediction on a large-scale synthetic flying chair dataset. However, the results of FlowNet do not match the performance of DM on realistic datasets. This motivates us to reformulate DM~\cite{Revaud_ijcv2015} as an end-to-end trainable neural network.

Beyond CNNs, many authors have applied machine learning techniques to matching and optical flow. Sun \etal~\cite{Sun/eccv2008} investigate the statistical properties of optical flow and learn the regularizers using Gaussian scale mixtures, Rosenbaum \etal~\cite{Rosenbaum/nips2013} use Gaussian mixture models to model the statistics of optical flow, and Black~\etal.~\cite{Black/cvpr1997} apply the idea of principal components analysis to optical flow. Kennedy and Taylor~\cite{Kennedy/emmcvpr2015} train classifiers to choose different inertial estimatiors for optical flow. Leordeanu~\etal~\cite{Leordeanu/iccv2013} obtain occlusion probabilities by learning classifiers. Menze~\etal~\cite{Menze/gcpr2015} formulate optical flow estimation as a discrete inference problem in a conditional random field, followed by sub-pixel refinement. In these works, tuning feature parameters is mostly done separately and manually. In contrast to these works, our work aims to convert the whole quasi-dense matching pipeline into an end-to-end trainable CNN.

%% file: approach.tex
\section{Method}\label{s:method}

\begin{figure}[t]
\begin{center}
\includegraphics[width=\textwidth]{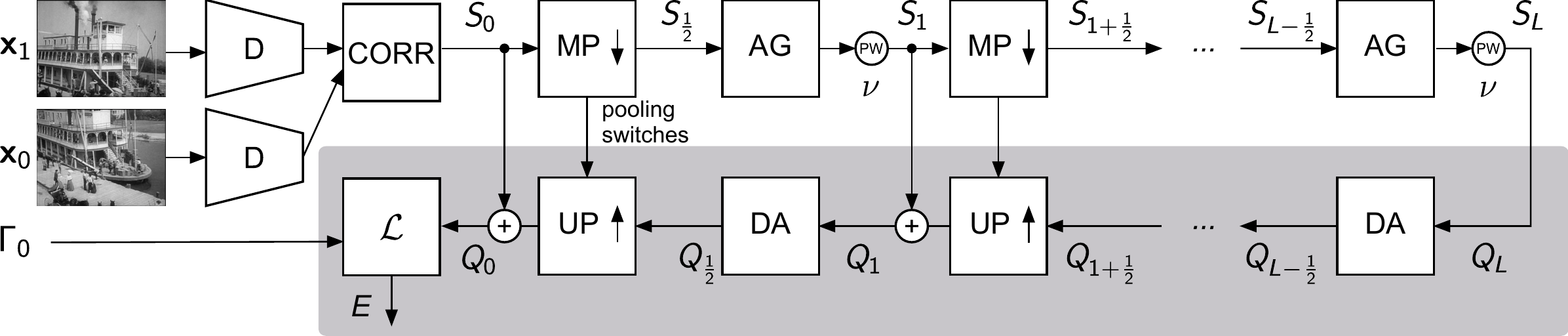}
\end{center}
\caption{End-to-end deep matching architecture, involving the following layers: descriptor extraction (D), correlation (CORR), max pooling (MP), aggregation (AG), power (PW), disaggregation (DA), unpooling (UP), summation (+), and structured loss ($\mathcal{L} $). The shaded area encloses our contribution, which amounts: formulating the DM decoding algorithm as a sequence of convolutional neural network layers supporting backpropagation.}\label{f:arch}
\end{figure}

Our key contribution is to show that the full DM pipeline can be formulated as a CNN with a $U$-topology (Fig.~\ref{f:arch}). The fine-to-coarse stage of DM was already given as a CNN in~\cite{Revaud_ijcv2015}. Here, we complete the construction and show that the DM recursive decoding stage can: (1) be interpreted as dynamic programming and (2) be implemented by convolutional operators which reverse the ones used in the fine-to-coarse stage (Sect.~\ref{s:endtoend}). The architecture can be trained using backpropagation, for which we propose a structured-output loss (Sect.~\ref{s:training}).

\subsection{Fully-Trainable Deep Matching Architecture}\label{s:endtoend}

In this section we formulate the \emph{complete} DM algorithm as a CNN. Consider a \emph{reference image} $I_0(\bp), \bp=(p_1,p_2)$ and a \emph{target image} $I_1(\bq), \bq = (q_1,q_2)$. The goal is to estimate a \emph{correspondence field} $\Gamma : \mathbb{R}^2\rightarrow\mathbb{R}^2,\bp \mapsto \bq$ mapping points $\bp$ in the reference image to corresponding points $\bq$ in the target image. The correspondence field is found as the maximizer
\begin{equation}\label{e:scoring}
\Gamma(\bp) = \operatornamewithlimits{argmax}_{\bq} S_0(\bq|\bp)
\end{equation}
of a scoring function $S_0(\bq|\bp)$ that encodes the similarity of point $\bp$ in $I_0$ with point $\bq$ in $I_1$ (the score has of course an implicit dependency on $I_0$ and $I_1$).\footnote{As proposed in DM, matches can be verified by testing whether they maximize the score also when going from the target image $I_1$ back to the reference image image $I_0$:
$
\operatorname{verified}(\bp) = [  \forall \bp' : Q(\Gamma(\bp) | \bp) \geq Q(\Gamma(\bp) | \bp')].
$}

A simple way of defining the scoring function $S_0$ is to compare patch descriptors. Thus, let $\phi(\bq|I) \in \mathbb{R}^d$ be a visual descriptor of a patch centred at $\bq$ in image $I$; furthermore, assume that $\phi$ is $L^2$ normalised. The score of the match $\bp \mapsto \bq$ can be defined as the cosine similarity of local descriptors, given by the inner product:
\begin{equation}\label{e:score-simple}
S_0(\bq|\bp) = \langle \phi(\bp|I_0), \phi(\bq|I_1) \rangle.
\end{equation}
A significant drawback of this scoring function is that it pools information only locally, from the compared patches. Therefore, unless all patches have a highly distinctive local appearance, many of the matches established by eq.~\eqref{e:scoring} are likely to be incorrect. 

Correcting these errors requires integrating global information in the score maps. In order to do so, DM builds a sequence of scoring functions $S_l(\bq|\bp), l = 0,1,2,\dots,L$ which are increasingly coarse but that incorporate information from increasingly larger image neighborhoods (Fig.~\ref{f:splash} top). Given these maps, equation \eqref{e:score-simple} is replaced by a recursive decoding process that extracts matches by analysing $S_L,S_{L-1},\dots,S_0$ in reverse order.

While the authors of~\cite{Revaud_ijcv2015} already showed that maps $S_l$ can be computed by convolutional operators, they did not formulate the decoding stage of DM as a network supporting end-to-end learning. Here we show that the recursive decoding process can be reformulated as the computation of additional score maps $Q_l(\bq|\bp), l = L, L-1, \dots, 1$ (Fig.~\ref{f:splash} bottom) by reversing the convolutional operators used to compute $S_0,S_1,\dots,S_L$. The two stages, fine to coarse and coarse to fine, are described in detail below.

\paragraph{Stage 1: Fine to coarse.}\label{s:aggr} DM starts with the scoring function $S_0$, computed by comparing local patches as explained above, and builds the other scores by alternating two operations: max pooling and aggregation.

The \textbf{max pooling} step pools scores $S_l$ with respect to the first argument $\bq$ in a square of side of $2^{l+1}\eta_0$ pixels, where $\eta_0$ is a parameter. This results in an intermediate scoring function $S_{l+1/2}$:
\begin{equation}\label{e:maxpool}
S_{l+\frac{1}{2}}(\bq|\bp) = 
\max
\left\{
 S_{l}(\bq'|\bp),\ \forall \bq':  \|\bq'-\bq\|_\infty \leq 2^l\eta_0
\right\}.
\end{equation}
In the following, the locations of the local maxima, also known as \emph{pooling switches}, will be denoted as $\bq'=m_{l}(\bq|\bp)$, where $m_l$ is defined such that
$
S_{l+1/2}(\bq|\bp) = S_{l}(m_{l}(\bq|\bp)|\bp).
$
Note that max pooling is exactly the same operator as commonly defined in convolutional neural networks. The resulting score $S_{l+1/2}(\bq|\bp)$ can be interpreted as the strength of the best match between $\bp$ in the reference image and all points within a distance $2^l \eta_0$ from $\bq$ in the target image.

After max pooling, the scores are \textbf{aggregated} at the four corners of a square patch of side $2^l\delta_0$ pixels:
\begin{equation}\label{e:aggr}
S_{l+1}(\bq|\bp) = 
\left[
\frac{1}{4} \sum_{i=1}^4
S_{l+\frac{1}{2}}(\bq + 2^l\bdelta_i| \bp + 2^l\bdelta_i)
\right]^\nu
\end{equation}
where $\bdelta_i = (\delta_0/2)\bepsilon_i$, $\delta_0 > 0$ is a parameter, and $\bepsilon_i$ are the unit displacement vectors:
$$
\bepsilon_1 = \begin{bmatrix} -1 \\ -1 \end{bmatrix}, \quad
\bepsilon_2 = \begin{bmatrix} -1 \\ +1 \end{bmatrix}, \quad
\bepsilon_3 = \begin{bmatrix} +1 \\ +1 \end{bmatrix}, \quad
\bepsilon_4 = \begin{bmatrix} +1 \\ -1 \end{bmatrix}.
$$
The exponent $\nu$ (set to 1.4 in DM) monotonically rescales the scores, emphasising larger ones. As detailed in~\cite{Revaud_ijcv2015}, the score $S_l(\bq|\bp)$ can be roughly interpreted as the likelihood that a deformable square patch of side $2^{l+1}\delta_0$ centered at $\bp$ in the reference image $I_0$ matches an analogous deformable patch centered at $\bq$ in the target image $I_1$.

Eq.~\eqref{e:aggr} can be rewritten as the convolution of $S_{l+1/2}$ with a particular 4D filter. Note that most neural network toolboxes are limited to 2+1D or 3+1D convolutions (with 2 or 3 spatial dimension plus one spanning feature channels), whereas here there are four spatial dimensions (given by the join of $\bp$ and $\bq$) and one feature channel, \ie the convolution is 4+1D. Hence, while implementing aggregation through convolution is more general, for the particular filter used in DM a direct implementation of \eqref{e:aggr} is much simpler.

\paragraph{Part 2: Coarse to fine.} In the original DM, scores $S_0,S_1,\dots,S_L$ are decoded by a recursive algorithm to obtain the final correspondence field. Here, we give an equivalent algorithm that uses only layer-wise and convolutional operators, with the major advantage of turning DM in an end-to-end learnable convolutional network. Another significant advantage is that the final product is a full, refined score map $Q_{0}$ assigning a confidence to all possible matches rather than finding only the best ones.

Since the last operation in the fist stage was to apply aggregation to $S_{L-1/2}$ to obtain $S_L$, the first operation in the reverse order is \textbf{disaggregation}. In general, $Q_{l+1}$ is disaggregated to obtain $Q_{l+1/2}$ as follows:
\begin{equation}\label{e:decode-aggr}
	Q_{l+\frac{1}{2}}(\bq|\bp)
	=
	\max
	\left\{ Q_{l+1}(\bq - 2^l \bdelta_i | \bp - 2^l \bdelta_i),\ i=1,2,3,4 \right\}.
\end{equation}
Disaggregation is similar to \emph{deconvolution}~\cite{Zeiler/cvpr2010,Long/cvpr2015,Noh/iccv2015,Ronneberger/miccai2015} or \emph{convolution transpose}~\cite{Vedaldi/mm2015} as it reverses a linear filtering operation. However, a key difference is that overlapping contributions are maxed out rather than summed.

Next, $Q_l$ is obtained by \textbf{unpooling} $Q_{l+1/2}$ and adding the result to $S_{l}(\bq|\bp)$:
\begin{equation}\label{e:decode-maxpool}
	Q_{l}(\bq|\bp)
	=
	S_{l}(\bq|\bp)
	+
	\max \left\{ Q_{l+\frac{1}{2}}(\bq'|\bp), \forall \bq' :  m_l(\bq'|\bp) = \bq \right\}\cup \{- \infty\}.
	\end{equation}
Unpooling is also found in architectures such as deconvnets; however here 1) the result is infilled with $-\infty$ rather than zeros and 2) overlapping unpooled values are maxed out rather than summed. The result of unpooling is summed to $S_{l}(\bq|\bp)$ to mix coarse and fine grained information.

Next, we discuss the equivalence of these operations to the original DM decoding algorithm. In the fine to coarse stage, through pooling and aggregation, the score $S_0(\bq_0|\bp_0)$ contributes to the formation of the coarser scores $S_l(\bq_1|\bp_1),\dots,S_L(\bq_L|\bp_L)$ along certain paths $(\bp_1,\bq_1), \dots, (\bp_L,\bq_L)$ restricted to the set:
\[
\mathcal{H}(\bq_0|\bp_0) =
\{
(\bp_0,\bq_0,\bp_1,\bq_1,\dots,\bq_L) :
\forall l\ \exists i:\
\bp_l = \bp_{l+1} - 2^l\delta_i,\ 
\bq_l = m_l(\bq_{l+1} - 2^l\delta_i|\bp_l)
\}.
\]
DM associates to the match $\bq_0|\bp_0$ the sum of the scores along the best of such paths:
\[
Q_0(\bq_0|\bp_0)
=
\max
\left\{
\sum_{l=0}^L
S_l(\bq_l|\bp_l):
\
(\bp_0,\bq_0,\bp_1,\bq_1,\dots,\bq_L) \in \mathcal{H}(\bq_0|\bp_0)
\right\}.
\]
DM uses recursion and memoization to compute this maximum efficiently; the disaggregation and unpooling steps given above implement a dynamic programming equivalent of this recursive algorithm. This is easily proved; empirically, the two implementations were found to be numerically equivalent as expected.

%
%
%
%

\subsection{Training and loss functions}\label{s:training}

Training with DM requires to define a suitable \emph{loss function} for the computed scoring function $S$. One possibility is to minimise the distance $\mathcal{L}(S,\Gamma_0)=\operatorname{mean}_{\bp,\bq}\|S(\bq|\bp) - g_\sigma(\bq -\Gamma_0(\bp))\|^2$ between $S$ and a smoothed indicator function $g_\sigma(z) = \exp(-\|z\|^2/2\sigma^2)$ of the ground truth correspondence field $\Gamma_0$. While a similar loss is often used to learn keypoint detectors with neural networks~\cite{Long/cvpr2014,Han/cvpr2015}, it has two drawbacks: first, it requires scores to attain pre-specified values when only relative values are relevant and, second, the loss must be carefully rebalanced as $g_\sigma(\bq -\Gamma_0(\bp)) \approx 0$ for the vast majority of pairs $(\bp,\bq)$.

In order to avoid these issues, we propose to use instead the following \emph{structured output} loss:
\[
\mathcal{L}(S,\Gamma_0)
=
\sum_{\bp,\bq} \max\{0, 1 - g_\sigma(\bq -\Gamma_0(\bp)) + S(\bq|\bp) - S(\Gamma_0(\bp)|\bp) \}.
\]
Here, the term $1-g_\sigma(\bq -\Gamma_0(\bp))$ defines a variable margin for the hinge loss, small when $\bq \approx \Gamma_0(\bp)$ and close to 1 otherwise. This loss looks at relative scores; in fact $\mathcal{L}(S,\Gamma_0)=0$ requires the correct matches to have score larger than incorrect ones. Furthermore, it is automatically balanced as each term in the summation involves comparing the score of a correct and an incorrect match.

Note that DM defines a whole hierarchy of score maps ($S_0,\dots,S_L,Q_L,\dots,Q_0$) and a loss can be applied to each level of the hierarchy. In general, we expect application at the last level $Q_L$ to be the most important, as this reflects the final output of the algorithm, but combinations are possible. For $n$ training image pairs $(\bx^{(i)}_0,\bx^{(i)}_1, \Gamma^{(i)})$, and by denoting with $\bw$ the parameters of DM, learning reduces to optimizing the objective function:
\[
 \min_{\bw}
 \frac{\lambda}{2} \|\bw\|^2
 +
 \frac{1}{n}
 \sum_{i=1}^n
 \mathcal{L}(Q_L(\bx^{(i)}_0,\bx^{(i)}_1;\bw), \Gamma^{(i)}_0).
\] 
We follow the standard approach of optimizing the objective using (stochastic) gradient descent~\cite{LeCun/ieee1998}. This requires computing the derivative of the loss and DM function $Q_L$ w.r.t. the parameters $\bw$, which can be done using backpropagation. Note that, while derivations are omitted, all layers in the DM architecture are amenable to backpropagation in the usual way.

%
%

\subsection{Discretization}\label{s:discretized}

So far, variables $\bq$ and $\bp$ have been treated as continuous. However, in a practical implementation these are discretized. By choosing a discretization scheme smartly, we can make the implementation more efficient and simpler. We describe such a scheme here.

For efficiency, DM doubles at each layer the sampling stride of the variable $\bq$ and restricts the match $\bq$ to be within a given maximum distance of $\bp$. Hence, $\bq$ is sampled as:
\[
\bq = 2^l\gamma_0 (\bk_{l} - 1 - R_l) + \bp,\quad
\bk_l \in \{1,\dots,2R_l+1\}^2,
\]
where $\bk_l$ is a discrete index, $\gamma_0$ is the sampling stride (in pixels) at level $l=0$, $\gamma_0 R_0$ the distance to $\bp$ at level 0, and
$
 R_{l+1} = \left\lceil {R_l} /{2} \right\rceil
$
is halved with each layer. In this expression, and in the rest of the section, summing a scalar to a vector means adding it to all its components.

For efficiency, DM is usually restricted to a quasi-dense grid of points $\bp$ in the reference image, given by:
\[
\bp = 
\alpha_0 
\left(
\bi_l 
- 1
+ \tau_l
\right) + \beta_0,
\quad
\bi_l \in \{1,\dots,H_l\}\times\{1,\dots,W_l\},
\quad
\tau_l = \mathbf{1}_{l\geq 1} 
\left(
\frac{\delta_0}{2\alpha_{0}} - \left\lceil \frac{\delta_0}{2\alpha_{0}} \right\rceil
\right).
\]
The parameters $\alpha_0$ and $\beta_0$ are the stride and offset of the patch descriptors extracted from the reference image and they remain constant at all layers; however, there is an additional variable offset $\tau_l$ to compensate for the effect of discretization in aggregation, as explained below. Here, the symbol $\mathbf{1}_{l\geq 1}$ is one if the condition $l \geq 1$ is satisfied and zero otherwise.

From these definitions, the discretized score maps, denoted with a bar, are given by
$
\bar S_{l}(\bk_l|\bi_l) =
S_{l}(\bq|\bp),
$
$
\bar S_{l+1/2}(\bk_{l+1}|\bi_l) =
S_{l+1/2}(\bq|\bp),$
and similarly for $\bar Q_l$. 

Simplifications arise by assuming that $\gamma_0$ divides exactly the pooling window size $\eta_0$, that $\alpha_0$ divides $\delta_0$, and that $\gamma_0$ divides $\alpha_0$. Under these assumptions, $\bar S_{l+1/2}(\bk_{l+1}|\bi_l)$ is obtained from $\bar S_l(\bk_l|\bi_l)$ by applying the standard CNN max pooling operator with a pooling window size $W=1+2\eta_0	/\gamma_0$ and padding $P=\eta_0/\gamma_0 + 2 R_{l+1} - R_l$. Note in particular that $W$ is the same at all layers. Since usually $\eta_0 = \gamma_0$, this amounts to $3 \times 3$ pooling with a padding of zero or one pixels. The discretized aggregation operator is also simple and given by:
\[
\bar S_{l+1}(\bk_{l+1}|\bi_{l+1}) 
= \frac{1}{4} \sum_{i=1}^4
\bar S_{l+\frac{1}{2}}\left(\bk_{l+1}\middle\rvert\bi_{l+1} + 2^l
\frac{\delta_0}{2\alpha_0}\bepsilon_i - \tau_1 \mathbf{1}_{l=0}\right).
\]
Note that, since $\bq$ is expressed relatively to $\bp$, aggregation reduces to averaging selected slices of the discretized score maps (\ie there is no shift applied to $\bk_{l+1}$). Note also that for $l \geq 1$, given that $\alpha_0$ divides $\delta_0$, the increment applied to the index $\bi_{l+1}$ is integer as required. For $l = 0$ and $\alpha_0 = \delta_0$ (as it is usually the case), the shift $\delta_0/2\alpha_0=1/2$ is fractional. In this case, however, the additional offset $\tau_1 = -1/2$ restores integer coordinates as needed.

%% file: experiments2.tex
\section{Experiments}\label{s:experiments}


The primary goal of this section is to demonstrate the benefit of learning the DM parameters using backpropagation compared to hand-tuning. There are several implementations of DM available online; we base ours on the GPU-based version by the original authors\footnote{\url{http://lear.inrialpes.fr/src/deepmatching/}.}~\cite{Revaud_ijcv2015}, except for the decoding stage for which we use their CPU version with memoization removed. We do so because this eliminats a few small approximations found in the original code. This version is the closest, and in fact numerically equivalent, to our implementation using MatConvNet~\cite{Vedaldi/mm2015} and our new convolutional operators.

\paragraph{Datasets.} The \textbf{MPI Sintel~\cite{Butler/ECCV2012}} dataset contains 1,041 image pairs and correspondence fields obtained from synthetic data (computer graphics). Scenes are carefully engineered to contain challenging conditions. There are two versions: clean and final (with effects such as motion blur and fog). We consider a subset of the Sintel clean training set to evaluate our methodology. This is dubbed SintelMini, and consists of 7 sequences (313 images) for training and every 10th frame from a different set of 5 sequences (25 images) for validation.
The \textbf{FlyingChair dataset} by Fischer~\etal~\cite{Fischer/iccv2015} contains synthetically-generated data as Sintel, but with abstract scenes consisting of ``flying chairs''. It consists of respectively 22,232/640 train/val image pairs and corresponding flow fields. These images are generated by rendering 3D chair models in front of random background images from Flickr, while the motions of both the chairs and the background are purely planar.
The \textbf{KITTI flow 2012~\cite{Geiger/CVPR2012,Menze/CVPR2015}} dataset contains 194/195  training/testing image pairs and correspondence fields for road scenes. The data contains large baselines but only motions arising from driving a car. Ground truth correspondences are obtained using 3D laser scanner and hence are not available at all pixels. Furthermore, the flow is improved by fitting 3D CAD models to observed vehicles on the road and using those to compute displacements.%
%

\paragraph{Evaluation metrics.} In order to measure matching accuracy, we adopt the \textbf{accuracy@T} metric of Revaud~\etal~\cite{Revaud_ijcv2015}. Given the ground truth and estimated dense correspondence fields $\Gamma_0, \Gamma:\bp \mapsto \bq$ from image $I_0:\Omega_0\rightarrow \mathbb{R}$ to image $I_1:\Omega_1\rightarrow\mathbb{R}$, accuracy@T is the fraction of pixels in $\Omega_0$ correctly matched up to an error of $T$ pixels, \ie $|\{\bq \in \Omega_0 : \|\Gamma_0(\bq) - \Gamma_0(\bq)\|\leq T\}|/|\Omega_0|$.\footnote{Following~\cite{Revaud_ijcv2015}, the quasi-dense DM matches are first filtered by reciprocal verification and then correspondences are propagated to all pixels by assigning to each point $\bq$ the same displacement vector of the most confident available nearest available neighbor $\bq'$ within a $L^\infty$-radius of 8 pixels by setting $\Gamma(\bq) = \Gamma(\bq') - \bq' + \bq$.} In addition to accuracy@T, we also consider the \textbf{end point error} (EPE), obtained as the average correspondence error $\operatorname{mean}_{\bq \in \Omega_0} \|\Gamma(\bq) - \Gamma_0(\bq)\|$. In all cases, scores are averaged over all image pairs to yield the final result for a given dataset. If ground truth correspondences are available only at a subset of image locations, $\Omega_0$ is restricted to this set in the definitions above. 
For the KITTI dataset, we report in particular results for $\Omega_0$ restricted to non-occluded ares (NOC) and all areas (OCC).

\paragraph{Implementation details.} For DM, unless otherwise stated we use $L=6$ layers, $R=80$ pixels, $\alpha_0=\delta=8$, $\beta_0=4$, $\gamma_0=1$, $\eta_0=1.4$. Training uses an NVIDIA Titan X GPU with 12 GBs of on-board memory. Training uses stochastic gradient descent with momentum with mini-batches comprising one image pair at a time (note that an image pair can be seen as the equivalent of a very large batch of image patches).

\subsection{Results}\label{s:results}

\newcommand{\nocheck}{$\times$}

\begin{table}[t]
\footnotesize{
\begin{center}
\begin{tabular}{cll|cc|cccccc}
\hline

& Patch    & Training set & \multicolumn{2}{c|}{Elements learned}   &  Acc@2  &  Acc@5  &  Acc@10  &  EPE        &  EPE     \\
& descr.   &              & expon.     & features                   &         &         &          &  (matches)  &  (flow)   \\   
\hline
\hline

(a) & HOG      & \textemdash   & \nocheck   & \nocheck & 84.52\%  &  91.89\%  &  94.36\%  &  3.83  &  1.88     \\
(b) & HOG      & Sintel Mini   & \checkmark   & \nocheck & 84.59\%  &  92.03\%  &  94.49\%  &  3.73  &  1.84  \\
(c) & CNN      & \textemdash   & \nocheck   & \nocheck & 85.28\%  &  92.25\%  &  94.83\%  &  3.58  &  1.80   \\
(d) & CNN      & Sintel Mini    & \checkmark   & \nocheck & 85.30\%  &  92.27\%  &  94.87\%  &  3.70  &  1.64  \\
(e) & CNN      & Sintel Mini    & \nocheck   & \checkmark & \textbf{86.81\%}  &  92.52\%  &  94.86\%  &  3.37  &  1.60   \\
(f) & CNN      & Sintel Mini    & \checkmark    & \checkmark  & {86.79\%}  &  \textbf{92.58\%}  &  \textbf{94.90}\%  &  {3.34}  &  \textbf{1.57}  \\
(g) & CNN      & Flying Chairs  & \checkmark    & \checkmark  & 86.11\%  &  92.47\%  &  94.88\%  &  \textbf{3.33}  &  1.65   \\

\end{tabular}
\end{center}
}
\caption{\textbf{Fully-Trainable DM performance.} DM variants evaluated on Sintel Mini (see text) validation and trained on either Sintel Mini training or Flying Chairs. The top row corresponds to the baseline DM algorithm, equivalent to the GPU version of~\cite{Revaud/cvpr2015}.}
\label{t:main}
\end{table}

\begin{table}[t]
\scriptsize{
\begin{center}
\begin{tabular}{lllcccc|cc}
\hline
Method  & Training      & Test          &  Acc@2   &  Acc@5  &  Acc@10            &  EPE         &  EPE          &  Err-OCC    \\
        &               &               &          &         &                    &   (matches)  &   (flow)      &   (flow 3px)  \\\hline
FlowNet S+v \cite{Fischer/iccv2015} & Flying Chairs &KITTI12&-               &-               &-               &-       &  6.50  &- \\
DM-HOG  &  \textemdash  & KITTI12 & 60.50\%  &  79.34\%&  84.27\%           &  11.39       &  \textbf{3.59}&  {16.56\%} \\
DM-CNN  &  \textemdash  & KITTI12 & 61.21\%  &  78.81\%&  84.01\%           &  12.29       &  4.11         &  17.78\% \\
DM-CNN  &  Flying Chairs& KITTI12 & \textbf{63.90\%}  &  \textbf{80.11\%}  &  \textbf{84.71\%}  &  \textbf{11.12}  &  3.61  &  \textbf{16.41\%}\\
\hline
FlowNet S+v \cite{Fischer/iccv2015} & Flying Chairs &Sintel Final&-               &-               &-                 &-       &  4.76  &- \\
DM \cite{Revaud_ijcv2015}      &\textemdash    &Sintel Final&-               &-               &  89.2\%          &-       &   4.10 &- \\
DM-HOG                      &\textemdash    &Sintel Final& 74.37\%        &  85.26\%       &  89.39\%         &  7.08  &  3.72  &  11.44\% \\
DM-CNN                      &\textemdash    &Sintel Final& 75.15\%        &  85.42\%       &  89.48\%         &  7.03  &  3.63  &  11.52\% \\
DM-CNN                      & Flying Chairs &Sintel Final&\textbf{76.55\%}&\textbf{86.22\%}&\textbf{90.03\%}  &  \textbf{6.77}  &  \textbf{3.50}  &  \textbf{11.10\%} \\

\hline
FlowNet C+v \cite{Fischer/iccv2015} & Flying Chairs &Sintel Clean&-               &-               &-               &-            &  3.57       &-\\
DM-HOG                      &\textemdash    &Sintel Clean& 82.51\%        &  90.18\%       &  92.70\%       &  5.26       &  2.32       &  7.00\% \\
DM-CNN                      &\textemdash    &Sintel Clean& 83.03\%        &  90.24\%       &  92.87\%       &  5.22       &  2.25       &  6.85\% \\
DM-CNN                      &Flying Chairs  &Sintel Clean&\textbf{84.16\%}&\textbf{90.85\%}&\textbf{93.31\%}&\textbf{4.78}&\textbf{2.14}&  \textbf{6.51\%} \\

\end{tabular}
\end{center}
}
\caption{\textbf{Performance comparison.} We train DM variants on large-scale synthetic dataset Flying Chairs, and evaluate on KITTI 12 train and Sintel train.
{Acc}@{$n$}~\cite{Revaud_ijcv2015} assigns each pixel a nearby match, measuring the proportion correct within $n$ pixels. EPE (endpoint error) is the mean euclidean distance between estimated flow vectors
and the ground truth (considering just pixels where ground truth is available). EPE (matches) is computed only at the positions where we have our quasi-dense matches. EPE (flow) measures the endpoint error for the flow estimation, where flow is produced by post-processing the matches with EpicFlow~\cite{Revaud/cvpr2015}. Err-OCC likewise measures the dense
flow, giving proportion of flows off by more than 3 pixels. The version excluding occlusions, Err-NOC, is given in the text.
}
\label{t:sec}
\end{table}

\paragraph{End-to-end DM training.} In our first experiment (Table~\ref{t:main}) we evaluate several variants of DM training. To do so, we consider the smaller and hence more efficient Sintel Mini dataset, a subset of Sintel described above. In Table~\ref{t:main} (a) vs (b) we compare using the default value of $\nu=1.4$ used to modulate the output of the aggregation layers and learning values $\nu_l,l=1,\dots,L$ specific for each layer. Even with this simple change there is a noticeable improvement (+0.13\% acc@10). Next, we replace the HOG features with a trainable CNN architecture $\phi$ to extract descriptors from image patches. We use the first four convolutional layers (conv1\_1, conv1\_2, conv2\_1, conv2\_2) of the pre-trained VGG-VD network~\cite{simonyan15very}. Just by replacing the features, we notice a further improvement ((a) vs (c) +0.47\% acc@10) of DM, which can be increased by learning the DM exponents (d). Most interestingly, in (f) we obtain a further improvement by back-propagating from DM to the feature extraction layers and optimizing the features themselves (hence achieving end-to-end training from the raw pixels to the matching result). The last experiment (g) shows that similar improvements can also be obtained by training from completely unrelated datasets, namely Flying Chairs, indicating that learning generalizes well.

\paragraph{Standard benchmark comparisons.} To test DM training in realistic scenarios, we evaluate  performance on two standard benchmarks, namely the Sintel and KITTI 2012 training sets (Table~\ref{t:sec}) as these have publicly-available ground truth to compute accuracy. For training, we use Flying Chairs,
which is designed to be statistically similar to the Sintel target dataset. Compared to the HOG-DM baseline, training the CNN patch descriptors in DM improves accuracy@10 by +0.44\% on KITTI and by +0.64\% on Sintel Final.

An application of DM is optical flow, where it is usually followed by interpolation and refinement such as Brox and Malik~\cite{Brox/pami2011} or EpicFlow~\cite{Revaud/cvpr2015}. We use EpicFlow to interpolate our quasi-dense matches and compare the EPE results of FlowNet~\cite{Fischer/iccv2015}. While there are better methods than FlowNet for optical flow estimation, we choose it for comparison as this was proposed as a fully-trainable CNN for dense image matching; we compare to their results using variational refinement (+v) which is similar to EpicFlow interpolation. We train our method on Flying Chairs to allow a direct comparison with the results reported in~\cite{Fischer/iccv2015}. 

Compared to the pretrained CNN, training further on Flying Chairs gives a notable improvement in EPE, decreasing from 3.63 to 3.50 for Sintel Final and from 4.11 to 3.61 for KITTI. Compared to HOG, the improvement is even greater for Sintel Final, a gap of 0.22 pixels, however for KITTI the CNN is initially worse than
HOG. Training on synthetic data improves most metrics on KITTI, with the exception of EPE (flow). We believe the latter result to be due to the fact that the EpicFlow refinement step, which is not trained, is not optimally tuned to the different statistics of the improved quasi-dense matches. The refinement step is in fact known to be sensitive to the data statistics (for example, in~\cite{Revaud/cvpr2015} different tunings are used for different datasets). If we exclude occlusions in the ground truth
for KITTI, our trained CNN gets EPE-NOC of 1.43 compared to 1.51 for HOG, and Err-NOC falls from 7.84\%  to 7.41\%.


FlowNet EPEs on KITTI12-Train and Sintel Final Train are respectively 6.50 and 4.76, whereas our trained DM-CNN model has EPEs of 3.61 and 3.50 respectively. This confirms the benefit of the DM architecture, which we turn into a CNN in this paper.

%
%